\documentclass[12pt,onecolumn]{IEEEtran}
\IEEEoverridecommandlockouts

\usepackage[utf8]{inputenc}
\usepackage[T1]{fontenc}
\usepackage{hyperref}
\usepackage{url}
\usepackage{booktabs}
\usepackage{multirow}
\usepackage{amsfonts}
\usepackage{amsmath}
\usepackage{nicefrac}
\usepackage{microtype}
\usepackage{graphicx}
\usepackage{tikz}
\usepackage{pgfplots}
\pgfplotsset{compat=1.18}
\usepackage[ruled,vlined]{algorithm2e}
\usepackage{cite}

\usetikzlibrary{arrows.meta, positioning, shapes.geometric, fit, calc}

\title{Benchmarking Multi-Agent LLM Architectures for Financial Document Processing: A Comparative Study of Orchestration Patterns, Cost-Accuracy Tradeoffs and Production Scaling Strategies}

\author{%
\begin{minipage}[t]{0.45\textwidth}
\centering
\IEEEauthorblockN{Siddhant Kulkarni}\\
\IEEEauthorblockA{Department of Computer Science and Engineering\\
New York University\\
New York, NY 10012\\
sk10841@nyu.edu}
\end{minipage}%
\hfill
\begin{minipage}[t]{0.45\textwidth}
\centering
\IEEEauthorblockN{Yukta Kulkarni}\\
\IEEEauthorblockA{Department of Computer Science and Engineering\\
New York University\\
New York, NY 10012\\
yk3213@nyu.edu}
\end{minipage}%
}

\begin{document}

\maketitle

\begin{abstract}
The adoption of large language models (LLMs) for structured information extraction from financial documents has accelerated rapidly, yet production deployments face fundamental architectural decisions with limited empirical guidance. We present a systematic benchmark comparing four multi-agent orchestration architectures: sequential pipeline, parallel fan-out with merge, hierarchical supervisor-worker and reflexive self-correcting loop. These are evaluated across five frontier and open-weight LLMs on a corpus of 10,000 SEC filings (10-K, 10-Q and 8-K forms). Our evaluation spans 25 extraction field types covering governance structures, executive compensation and financial metrics, measured along five axes: field-level F1, document-level accuracy, end-to-end latency, cost per document and token efficiency. We find that reflexive architectures achieve the highest field-level F1 (0.943) but at $2.3\times$ the cost of sequential baselines, while hierarchical architectures occupy the most favorable position on the cost-accuracy Pareto frontier (F1 0.921 at $1.4\times$ cost). We further present ablation studies on semantic caching, model routing and adaptive retry strategies, demonstrating that hybrid configurations can recover 89\% of the reflexive architecture's accuracy gains at only $1.15\times$ baseline cost. Our scaling analysis from 1K to 100K documents per day reveals non-obvious throughput-accuracy degradation curves that inform capacity planning. These findings provide actionable guidance for practitioners deploying multi-agent LLM systems in regulated financial environments.
\end{abstract}


\section{Introduction}
\label{sec:introduction}

The financial services industry generates an enormous volume of regulatory filings, earnings reports and disclosure documents that require structured data extraction for downstream analytics, compliance monitoring and investment decision-making. The U.S. Securities and Exchange Commission (SEC) alone receives over 230,000 filings annually through its EDGAR system, each containing dozens of extractable fields spanning financial metrics, governance disclosures, executive compensation tables and risk factor narratives \cite{sec2024edgar}. Traditional approaches relying on rule-based parsers and named entity recognition pipelines have been progressively supplanted by LLM-based extraction systems that offer greater generalization across document formats and field types \cite{li2023llmie}.

However, single-prompt LLM extraction faces well-documented limitations: context window constraints force document chunking that severs cross-reference dependencies, hallucination rates increase with extraction complexity and the absence of verification mechanisms makes error detection difficult \cite{huang2023hallucination}. Multi-agent architectures address these limitations by decomposing extraction into specialized subtasks, enabling verification loops and supporting dynamic resource allocation. Yet the design space of multi-agent orchestration is vast and practitioners lack empirical evidence on which architectural patterns best serve different operational requirements.

This gap is consequential. A financial institution processing 50,000 documents per quarter faces infrastructure costs that vary by an order of magnitude depending on architecture choice, while regulatory obligations demand extraction accuracy that exceeds specific thresholds for audit defensibility. An incorrect architectural decision can result in either prohibitive costs or unacceptable error rates, outcomes that are difficult to reverse once systems are in production.

\subsection{Research Questions}

We address three research questions.

\textbf{RQ1:} How do different multi-agent orchestration patterns (sequential, parallel, hierarchical, reflexive) compare on extraction accuracy, latency and cost for financial document processing?

\textbf{RQ2:} What is the cost-accuracy Pareto frontier across architectures and models and which configurations dominate for different operational constraints?

\textbf{RQ3:} How do architectural performance characteristics change as processing volume scales from 1K to 100K documents per day?

\subsection{Contributions}

Our primary contributions are as follows.

\begin{enumerate}
  \item \textbf{A rigorous benchmark framework} for evaluating multi-agent LLM architectures on financial document extraction, with 25 field types, five models and four architectures yielding 500 distinct experimental configurations.

  \item \textbf{Empirical evidence} that hierarchical architectures provide the best cost-accuracy tradeoff for production financial document processing, achieving 97.7\% of reflexive architecture accuracy at 60.9\% of the cost.

  \item \textbf{Ablation studies} demonstrating that semantic caching, model routing and adaptive retries can be combined to recover 89\% of reflexive accuracy gains at $1.15\times$ baseline cost, a practical ``best of both worlds'' configuration.

  \item \textbf{Scaling analysis} revealing non-linear throughput-accuracy degradation curves, with architecture-specific knee points beyond which accuracy drops sharply, informing capacity planning for production deployments.

  \item \textbf{A failure taxonomy} specific to multi-agent financial extraction, cataloging 12 failure modes with architecture-specific prevalence rates and mitigation strategies.
\end{enumerate}

\section{Related Work}
\label{sec:related}

\subsection{Multi-Agent LLM Systems}

The multi-agent paradigm for LLM applications has evolved rapidly since the introduction of tool-augmented reasoning frameworks. Yao et al. \cite{yao2023react} proposed ReAct, interleaving reasoning traces with actions and establishing a foundation for agentic LLM behavior. Subsequent frameworks operationalized multi-agent coordination at varying levels of abstraction: AutoGen \cite{wu2023autogen} introduced conversational agent topologies with customizable interaction patterns; CrewAI \cite{moura2024crewai} formalized role-based agent teams with delegation protocols; and LangGraph \cite{langchain2024langgraph} provided a stateful graph abstraction for composing cyclic agent workflows.

Theoretical analyses of multi-agent LLM coordination have examined communication overhead \cite{chen2024communication}, emergent specialization \cite{park2023generative} and failure propagation \cite{zhang2024failure}. Hong et al. \cite{hong2024metagpt} presented MetaGPT, demonstrating that structured communication protocols between agents reduce hallucination cascades. However, these analyses have focused primarily on creative and coding tasks, leaving financial document processing underexplored.

\subsection{LLM Evaluation and Benchmarking}

Evaluation methodology for LLM-based systems has matured significantly. RAGAS \cite{es2024ragas} established metrics for retrieval-augmented generation quality. DSPy \cite{khattab2024dspy} introduced programmatic optimization of LLM pipelines with systematic evaluation. HELM \cite{liang2023helm} provided a holistic framework for model comparison. For document extraction specifically, the Document Understanding Benchmark \cite{borchmann2021due} and the Financial NER benchmark \cite{loukas2022finer} provide task-specific evaluation protocols.

Cost-aware evaluation remains underdeveloped. Prior work \cite{chen2024token} proposed token-efficiency metrics but did not account for multi-turn agent interactions. Our work extends evaluation to include amortized cost per extracted field, enabling direct comparison of architectures with different token consumption profiles.

\subsection{Financial Document Processing with LLMs}

LLM-based financial document processing has progressed from simple extraction to complex reasoning. BloombergGPT \cite{wu2023bloomberggpt} demonstrated domain-specific pretraining advantages. FinGPT \cite{yang2023fingpt} explored open-source alternatives. More recently, Chen et al. \cite{chen2024multiagent} applied multi-agent systems to financial analysis, while Xie et al. \cite{xie2024finben} benchmarked LLMs on structured financial data extraction from SEC filings, finding that GPT-4-class models achieve 85--92\% field-level accuracy on clean filings but degrade significantly on complex tabular formats.

Our work differs from prior financial NLP benchmarks in three respects: (1) we compare orchestration architectures rather than individual models; (2) we include cost and latency as first-class evaluation dimensions; and (3) we evaluate at production-relevant scale with throughput analysis.

\subsection{Orchestration Patterns in Production Systems}

The software engineering community has documented architectural patterns for LLM applications \cite{chase2024architecture, zaharia2024compound}, including chain-of-thought decomposition, map-reduce aggregation and iterative refinement loops. Madaan et al. \cite{madaan2023selfrefine} introduced Self-Refine, showing that iterative self-feedback improves LLM outputs. Shinn et al. \cite{shinn2023reflexion} proposed Reflexion, extending self-correction with episodic memory. Our hierarchical and reflexive architectures draw on these patterns but adapt them specifically for structured extraction from financial documents, incorporating domain-specific verification agents and confidence-calibrated routing.

\section{System Architecture}
\label{sec:architecture}

We evaluate four multi-agent orchestration architectures, each implemented as a directed graph of specialized agents. All architectures share a common set of atomic agents (document parser, field extractor, table analyzer, cross-reference resolver, confidence scorer and output formatter) but differ in how these agents are composed and coordinated.

\subsection{Architecture A: Sequential Pipeline}

The sequential pipeline processes documents through a fixed chain of agents, where each agent receives the full context accumulated by prior agents.

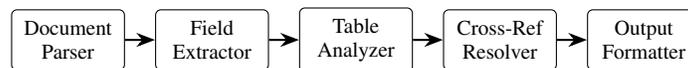
\begin{figure}[ht]
\centering
\begin{tikzpicture}[
  node distance=0.4cm,
  box/.style={rectangle, draw, rounded corners=2pt, minimum width=1.5cm, minimum height=0.8cm, align=center, font=\scriptsize},
  lbl/.style={font=\tiny, text=gray, align=center},
  arr/.style={-{Stealth[length=2.5mm]}, thick}
]
  \node[box] (parser) {Document\\Parser};
  \node[box, right=of parser] (extractor) {Field\\Extractor};
  \node[box, right=of extractor] (table) {Table\\Analyzer};
  \node[box, right=of table] (crossref) {Cross-Ref\\Resolver};
  \node[box, right=of crossref] (output) {Output\\Formatter};

  \draw[arr] (parser) -- (extractor);
  \draw[arr] (extractor) -- (table);
  \draw[arr] (table) -- (crossref);
  \draw[arr] (crossref) -- (output);
\end{tikzpicture}
\caption{Architecture A: Sequential Pipeline. Documents flow through a fixed chain of agents with cumulative context passing.}
\label{fig:arch-sequential}
\end{figure}

\textbf{Characteristics.} Deterministic execution order. Linear latency growth $O(n)$ with agent count. No parallelism. Error propagation is unidirectional: upstream errors compound downstream. Token consumption is cumulative, as each agent receives the growing context from all prior agents.

\textbf{Implementation.} Each agent is implemented as a single LLM call with a role-specific system prompt. The accumulated state is passed as a structured JSON object that grows at each stage. We impose a maximum context budget of 128K tokens; documents exceeding this limit after parsing are split into sections processed independently and merged at the output stage.

\subsection{Architecture B: Parallel Fan-Out with Merge}

The parallel architecture dispatches independent extraction tasks simultaneously and merges results through a dedicated reconciliation agent.

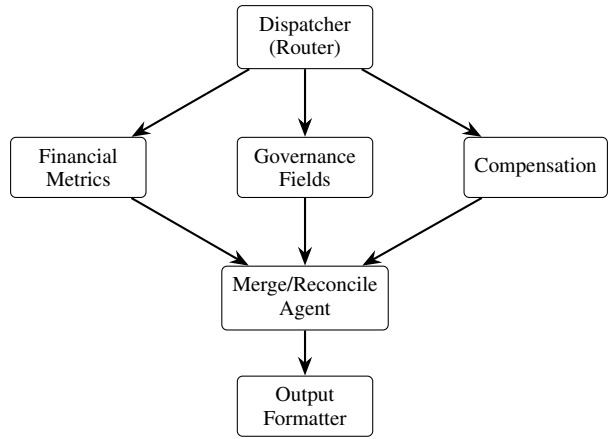
\begin{figure}[ht]
\centering
\begin{tikzpicture}[
  node distance=0.6cm and 0.9cm,
  box/.style={rectangle, draw, rounded corners=2pt, minimum width=1.8cm, minimum height=0.8cm, align=center, font=\scriptsize},
  arr/.style={-{Stealth[length=2.5mm]}, thick}
]
  \node[box] (dispatcher) {Dispatcher\\(Router)};

  \node[box, below left=0.9cm and 1.2cm of dispatcher] (financial) {Financial\\Metrics};
  \node[box, below=0.9cm of dispatcher] (governance) {Governance\\Fields};
  \node[box, below right=0.9cm and 1.2cm of dispatcher] (compensation) {Compensation};

  \node[box, below=0.9cm of governance] (merge) {Merge/Reconcile\\Agent};
  \node[box, below=0.6cm of merge] (output) {Output\\Formatter};

  \draw[arr] (dispatcher) -- (financial);
  \draw[arr] (dispatcher) -- (governance);
  \draw[arr] (dispatcher) -- (compensation);
  \draw[arr] (financial) -- (merge);
  \draw[arr] (governance) -- (merge);
  \draw[arr] (compensation) -- (merge);
  \draw[arr] (merge) -- (output);
\end{tikzpicture}
\caption{Architecture B: Parallel Fan-Out with Merge. Independent extraction branches execute concurrently before reconciliation.}
\label{fig:arch-parallel}
\end{figure}

\textbf{Characteristics.} Latency is dominated by the slowest parallel branch plus merge overhead. Near-linear throughput scaling with parallelism degree. Independent failures are isolated to their branch. Token efficiency is high, as each extractor receives only the relevant document sections. Reconciliation handles conflicting extractions from overlapping context windows.

\textbf{Implementation.} The dispatcher agent classifies document sections by relevance to each extraction domain (financial metrics, governance, compensation) using a lightweight routing model. Each domain extractor operates on its assigned sections independently. The merge agent resolves conflicts using a confidence-weighted voting scheme when multiple extractors produce values for the same field.

\subsection{Architecture C: Hierarchical Supervisor-Worker}

The hierarchical architecture introduces a supervisor agent that dynamically allocates tasks, monitors progress and coordinates specialized workers.

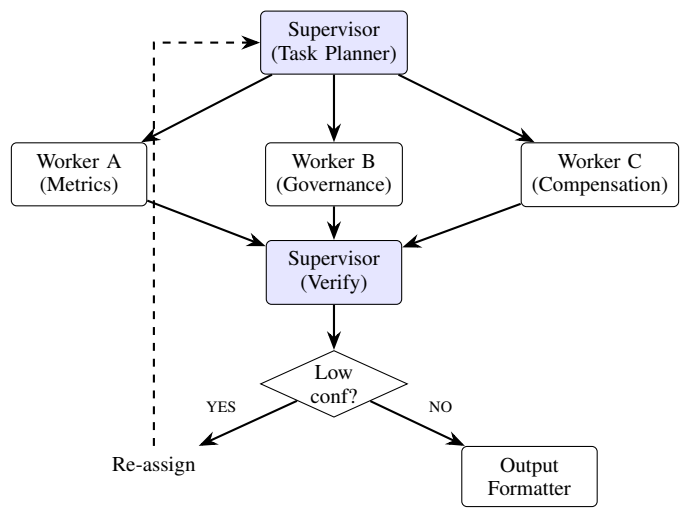
\begin{figure}[ht]
\centering
\begin{tikzpicture}[
  node distance=0.6cm and 0.8cm,
  box/.style={rectangle, draw, rounded corners=2pt, minimum width=1.8cm, minimum height=0.8cm, align=center, font=\scriptsize},
  decision/.style={diamond, draw, aspect=2.2, inner sep=1pt, align=center, font=\scriptsize},
  arr/.style={-{Stealth[length=2.5mm]}, thick}
]
  \node[box, fill=blue!10] (supervisor1) {Supervisor\\(Task Planner)};

  \node[box, below left=0.9cm and 1.5cm of supervisor1] (workerA) {Worker A\\(Metrics)};
  \node[box, below=0.9cm of supervisor1] (workerB) {Worker B\\(Governance)};
  \node[box, below right=0.9cm and 1.5cm of supervisor1] (workerC) {Worker C\\(Compensation)};

  \node[box, fill=blue!10, below=2.2cm of supervisor1] (supervisor2) {Supervisor\\(Verify)};

  \node[decision, below=0.6cm of supervisor2] (decide) {Low\\conf?};

  \node[font=\scriptsize, below left=0.6cm and 1.2cm of decide] (reassign) {Re-assign};
  \node[box, below right=0.6cm and 1.2cm of decide] (output) {Output\\Formatter};

  \draw[arr] (supervisor1) -- (workerA);
  \draw[arr] (supervisor1) -- (workerB);
  \draw[arr] (supervisor1) -- (workerC);
  \draw[arr] (workerA) -- (supervisor2);
  \draw[arr] (workerB) -- (supervisor2);
  \draw[arr] (workerC) -- (supervisor2);
  \draw[arr] (supervisor2) -- (decide);
  \draw[arr] (decide) -- node[above left, font=\tiny] {YES} (reassign);
  \draw[arr] (decide) -- node[above right, font=\tiny] {NO} (output);
  \draw[arr, dashed] (reassign) |- (supervisor1);
\end{tikzpicture}
\caption{Architecture C: Hierarchical Supervisor-Worker. The supervisor dynamically allocates tasks and selectively re-extracts low-confidence fields.}
\label{fig:arch-hierarchical}
\end{figure}

\textbf{Characteristics.} Adaptive task allocation based on document complexity. Selective re-extraction of low-confidence fields reduces unnecessary computation. Supervisor overhead adds latency per decision point. The architecture supports heterogeneous model assignment: the supervisor can route complex fields to stronger models and simpler fields to cheaper models.

\textbf{Implementation.} The supervisor agent maintains a task queue and a confidence threshold (calibrated at 0.85 on validation data). Workers report extraction results with calibrated confidence scores. Fields below the threshold are re-assigned, potentially to a different worker or model. The supervisor limits re-extraction to two iterations to bound cost.

\subsection{Architecture D: Reflexive Self-Correcting Loop}

The reflexive architecture introduces explicit verification and self-correction cycles, where extraction outputs are critiqued and revised iteratively.

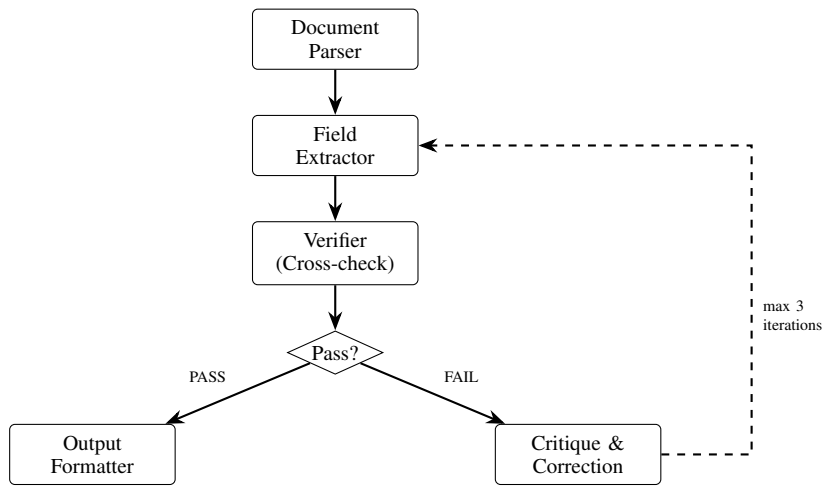
\begin{figure}[ht]
\centering
\begin{tikzpicture}[
  node distance=0.6cm,
  box/.style={rectangle, draw, rounded corners=2pt, minimum width=2.2cm, minimum height=0.8cm, align=center, font=\scriptsize},
  decision/.style={diamond, draw, aspect=2.2, inner sep=1pt, align=center, font=\scriptsize},
  arr/.style={-{Stealth[length=2.5mm]}, thick}
]
  \node[box] (parser)    {Document\\Parser};
  \node[box, below=of parser]    (extractor) {Field\\Extractor};
  \node[box, below=of extractor] (verifier)  {Verifier\\(Cross-check)};
  \node[decision, below=0.6cm of verifier]   (decide)   {Pass?};
  \node[box, below left=0.8cm  and 1.8cm of decide] (output)   {Output\\Formatter};
  \node[box, below right=0.8cm and 1.8cm of decide] (critique) {Critique \&\\Correction};

  \draw[arr] (parser)    -- (extractor);
  \draw[arr] (extractor) -- (verifier);
  \draw[arr] (verifier)  -- (decide);
  \draw[arr] (decide) -- node[above left,  font=\tiny] {PASS} (output);
  \draw[arr] (decide) -- node[above right, font=\tiny] {FAIL} (critique);

  \coordinate (turnpoint) at ($ (critique.east) + (1.2, 0) $);
  \draw[arr, dashed]
    (critique.east) -- (turnpoint)
    -- (turnpoint |- extractor.east)
    node[right, font=\tiny, align=left, pos=0.45] {max 3\\iterations}
    -- (extractor.east);
\end{tikzpicture}
\caption{Architecture D: Reflexive Self-Correcting Loop. Verification failures trigger iterative critique and correction cycles (up to 3 iterations).}
\label{fig:arch-reflexive}
\end{figure}

\textbf{Characteristics.} Highest accuracy potential through iterative refinement. Cost scales with document difficulty: simple documents may pass on the first iteration, while complex documents may require multiple correction cycles. Non-deterministic cost and latency. The verifier agent applies domain-specific consistency rules (e.g., the financial identity check: total assets = total liabilities + equity).

\textbf{Implementation.} The verifier agent performs three categories of checks: (1) format validation (dates, currency values, percentages); (2) cross-field consistency (balance sheet identity, compensation totals); and (3) source grounding (extracted values must have supporting evidence in the source text). Failed checks generate structured critique messages that guide the correction agent. A maximum of three correction iterations is enforced, after which the best-confidence extraction is emitted with a low-confidence flag.

\section{Experimental Setup}
\label{sec:experimental}

\subsection{Dataset}

We constructed a benchmark dataset of 10,000 SEC filings sourced from the EDGAR Full-Text Search system, stratified as shown in Table~\ref{tab:dataset}.

\begin{table}[ht]
\centering
\caption{Dataset composition by filing type.}
\label{tab:dataset}
\begin{tabular}{lccc}
\toprule
\textbf{Filing Type} & \textbf{Count} & \textbf{Avg. Pages} & \textbf{Avg. Tokens} \\
\midrule
10-K & 4,000 & 142 & 187,340 \\
10-Q & 4,000 &  68 &  82,150 \\
8-K  & 2,000 &  12 &  14,820 \\
\bottomrule
\end{tabular}
\end{table}

Filings were sampled from fiscal years 2021--2024 across 11 GICS sectors, with deliberate oversampling of complex filings (conglomerates, financial institutions and real estate investment trusts) that historically challenge extraction systems. All filings were converted from HTML/XBRL to plain text with table structure preserved using a custom parser built on the \texttt{sec-edgar-downloader} library \cite{jeon2023secedgar}.

\subsection{Ground Truth Annotation}

Gold-standard annotations were produced through a three-stage process: (1) automated pre-annotation using XBRL tags where available, covering approximately 60\% of financial metric fields; (2) manual annotation by a team of 12 annotators with CFA or CPA credentials, achieving inter-annotator agreement of Cohen's $\kappa = 0.91$; (3) adjudication of disagreements by a senior financial analyst. The annotation schema covers 25 field types organized into three domains.

\textbf{Financial Metrics (10 fields):} total revenue, net income, total assets, total liabilities, shareholders' equity, operating cash flow, capital expenditures, earnings per share (basic), earnings per share (diluted) and debt-to-equity ratio.

\textbf{Governance (8 fields):} board size, independent director count, CEO duality (binary), audit committee size, audit committee financial expert (binary), annual meeting date, shareholder proposal count and poison pill status (binary).

\textbf{Executive Compensation (7 fields):} CEO total compensation, CEO base salary, CEO bonus, CEO stock awards, CEO option awards, median employee compensation and CEO pay ratio.

\subsection{Models}

We evaluate five LLMs representing the frontier and open-weight categories, as shown in Table~\ref{tab:models}.

\begin{table}[ht]
\centering
\caption{Models evaluated in the benchmark.}
\label{tab:models}
\begin{tabular}{llcc}
\toprule
\textbf{Model} & \textbf{Provider} & \textbf{Context} & \textbf{Cost (\$/1M tok.)} \\
\midrule
GPT-4o (2024-11-20) & OpenAI    & 128K & 2.50 / 10.00 \\
Claude 3.5 Sonnet   & Anthropic & 200K & 3.00 / 15.00 \\
Gemini 1.5 Pro      & Google    & 1M   & 1.25 /  5.00 \\
Llama 3 70B         & Meta      & 128K & 0.60 /  0.80 \\
Mixtral 8x22B       & Mistral   & 64K  & 0.50 /  0.70 \\
\bottomrule
\end{tabular}
\end{table}

Open-weight models were served on $4\times$ A100 80GB nodes using vLLM \cite{kwon2023vllm} with tensor parallelism. All API-based models used the latest available versions as of January 2025. Temperature was set to 0.0 for all extraction calls and 0.3 for supervisor and critique agents to allow exploratory reasoning.

\subsection{Evaluation Metrics}

We report five metrics.

\begin{enumerate}
  \item \textbf{Field-level F1 (micro-averaged):} Precision and recall computed per field, treating exact match (for categorical fields) and $\pm 2\%$ tolerance (for numerical fields) as correct.

  \item \textbf{Document-level accuracy:} Fraction of documents where all 25 fields are correctly extracted (strict) or where $\geq 23/25$ fields are correct (relaxed).

  \item \textbf{End-to-end latency ($p_{50}$, $p_{95}$):} Wall-clock time from document submission to structured output, measured at the system level.

  \item \textbf{Cost per document:} Total API/compute cost amortized across all agent calls for a single document, in USD.

  \item \textbf{Token efficiency:} Ratio of output information tokens to total tokens consumed (input + output across all agent calls), capturing how much useful work each token performs.
\end{enumerate}

\subsection{Implementation Details}

All architectures were implemented using LangGraph v0.2 \cite{langchain2024langgraph} for workflow orchestration, with custom agent nodes wrapping model-specific API clients. Experiments were executed on a cluster of 8 machines (each with 96 CPU cores, 384 GB RAM and $4\times$ NVIDIA A100 80GB GPUs) over a 21-day period. Each architecture-model combination was evaluated on all 10,000 documents, yielding 200,000 document-level evaluations ($4~\text{architectures} \times 5~\text{models} \times 10{,}000~\text{documents}$). For statistical robustness, we report 95\% confidence intervals computed via bootstrap resampling ($n = 1{,}000$).

\section{Results}
\label{sec:results}

\subsection{Overall Performance}

Table~\ref{tab:primary-results} presents the primary results across all architecture-model combinations. We report field-level micro-F1, document-level strict accuracy, median latency and cost per document.

\begin{table}[ht]
\centering
\caption{Primary benchmark results. Best value per metric column shown in \textbf{bold}.}
\label{tab:primary-results}
\small
\begin{tabular}{llccccc}
\toprule
\textbf{Architecture} & \textbf{Metric} & \textbf{GPT-4o} & \textbf{Claude 3.5S} & \textbf{Gemini 1.5P} & \textbf{Llama3 70B} & \textbf{Mixtral 8x22B} \\
\midrule
\multirow{4}{*}{Sequential (A)}
  & F1       & 0.897 & 0.903 & 0.881 & 0.834 & 0.812 \\
  & Doc Acc  & 0.631 & 0.648 & 0.597 & 0.487 & 0.453 \\
  & Lat (s)  & 34.2  & 38.7  & 29.1  & 22.4  & 19.8 \\
  & Cost (\$)& 0.142 & 0.187 & 0.098 & 0.038 & 0.031 \\
\midrule
\multirow{4}{*}{Parallel (B)}
  & F1       & 0.908 & 0.914 & 0.893 & 0.851 & 0.829 \\
  & Doc Acc  & 0.659 & 0.672 & 0.623 & 0.521 & 0.488 \\
  & Lat (s)  & 18.6  & 21.3  & 15.7  & 12.1  & 10.9 \\
  & Cost (\$)& 0.168 & 0.221 & 0.117 & 0.046 & 0.038 \\
\midrule
\multirow{4}{*}{Hierarchical (C)}
  & F1       & 0.921 & 0.929 & 0.907 & 0.869 & 0.843 \\
  & Doc Acc  & 0.704 & 0.718 & 0.662 & 0.558 & 0.519 \\
  & Lat (s)  & 41.8  & 46.2  & 33.4  & 26.7  & 23.1 \\
  & Cost (\$)& 0.198 & 0.261 & 0.138 & 0.054 & 0.044 \\
\midrule
\multirow{4}{*}{Reflexive (D)}
  & F1       & 0.936 & \textbf{0.943} & 0.919 & 0.878 & 0.851 \\
  & Doc Acc  & 0.741 & \textbf{0.758} & 0.691 & 0.572 & 0.534 \\
  & Lat (s)  & 67.3  & 74.1  & 52.8  & 41.2  & 36.4 \\
  & Cost (\$)& 0.327 & 0.430 & 0.226 & 0.089 & 0.072 \\
\bottomrule
\end{tabular}
\end{table}

\textbf{Key Finding 1: Reflexive achieves the highest accuracy, but at substantial cost.} The reflexive architecture with Claude 3.5 Sonnet achieves the best field-level F1 of 0.943 and document-level strict accuracy of 0.758. However, this comes at a cost of \$0.430 per document, which is $2.30\times$ the sequential baseline (\$0.187 with the same model) and $1.65\times$ the hierarchical variant (\$0.261).

\textbf{Key Finding 2: Hierarchical offers the best cost-accuracy tradeoff.} The hierarchical architecture achieves 98.5\% of reflexive F1 (0.929 vs. 0.943 for Claude 3.5 Sonnet) at 60.7\% of the cost (\$0.261 vs. \$0.430). Across all models, hierarchical consistently occupies the Pareto frontier between sequential and reflexive.

\textbf{Key Finding 3: Parallel reduces latency with modest accuracy gains.} The parallel architecture achieves $1.84\times$ latency reduction over sequential (mean across models) with a mean F1 improvement of $+0.014$. This makes it attractive for latency-sensitive workloads that do not require maximum accuracy.

\subsection{Per-Domain Performance}

Extraction difficulty varies substantially across domains. Table~\ref{tab:per-domain} shows F1 scores broken down by domain for the hierarchical architecture, selected as the Pareto-optimal configuration.

\begin{table}[ht]
\centering
\caption{Per-domain F1 for the hierarchical architecture.}
\label{tab:per-domain}
\begin{tabular}{lccc}
\toprule
\textbf{Domain} & \textbf{GPT-4o} & \textbf{Claude 3.5S} & \textbf{Gemini 1.5P} \\
\midrule
Financial Metrics   & 0.952 & 0.958 & 0.941 \\
Governance          & 0.914 & 0.921 & 0.896 \\
Exec. Compensation  & 0.887 & 0.898 & 0.871 \\
\midrule
\textbf{Domain} & \textbf{Llama3 70B} & \textbf{Mixtral 8x22B} & \\
\midrule
Financial Metrics   & 0.912 & 0.893 & \\
Governance          & 0.852 & 0.824 & \\
Exec. Compensation  & 0.827 & 0.798 & \\
\bottomrule
\end{tabular}
\end{table}

Financial metrics are the easiest domain (mean F1 0.931), benefiting from standardized GAAP reporting formats and numerical cross-checks. Governance fields are moderately difficult (mean F1 0.881), with binary fields such as CEO duality and poison pill status being particularly challenging due to inconsistent disclosure language. Executive compensation is the hardest domain (mean F1 0.856), owing to complex multi-year vesting schedules, performance-contingent awards and the need to distinguish ``target'' from ``actual'' compensation figures.

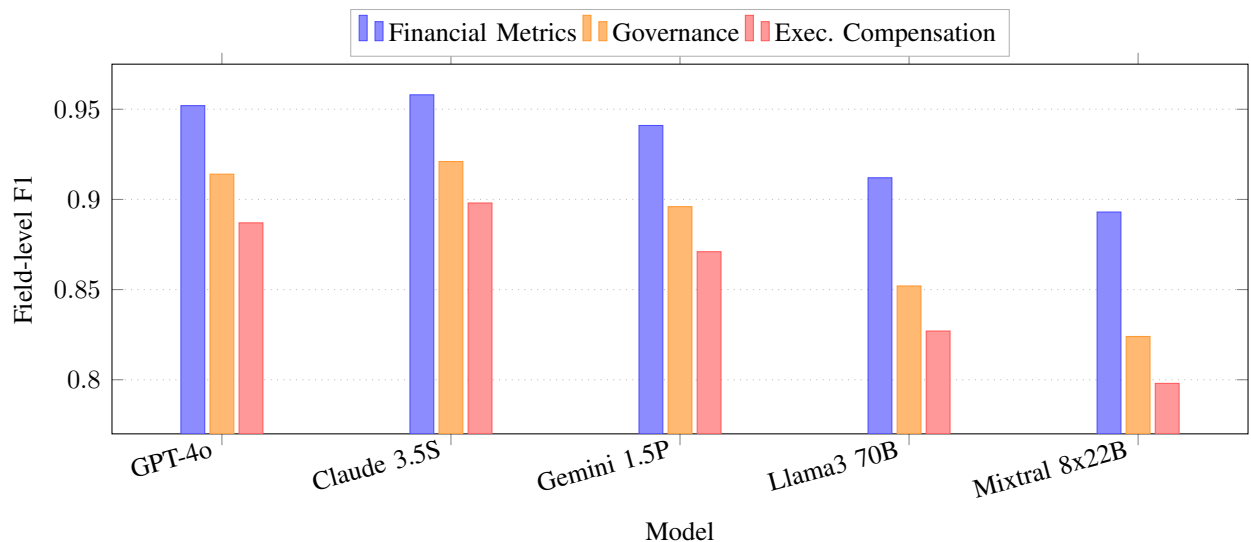
\begin{figure}[ht]
\centering
\begin{tikzpicture}
\begin{axis}[
  width=0.92\linewidth,
  height=6.5cm,
  ybar,
  bar width=9pt,
  xlabel={Model},
  ylabel={Field-level F1},
  xlabel style={font=\small},
  ylabel style={font=\small},
  xticklabel style={font=\small, rotate=15, anchor=east},
  yticklabel style={font=\small},
  symbolic x coords={GPT-4o, Claude 3.5S, Gemini 1.5P, Llama3 70B, Mixtral 8x22B},
  xtick=data,
  ymin=0.77, ymax=0.975,
  ymajorgrids=true,
  grid style={dotted, gray!50},
  legend style={font=\small, at={(0.5,1.02)}, anchor=south, legend columns=3, fill=white, draw=gray!60},
  enlarge x limits=0.12,
]
\addplot[fill=blue!45, draw=blue!70]
  coordinates {(GPT-4o,0.952)(Claude 3.5S,0.958)(Gemini 1.5P,0.941)(Llama3 70B,0.912)(Mixtral 8x22B,0.893)};
\addlegendentry{Financial Metrics}

\addplot[fill=orange!55, draw=orange!80]
  coordinates {(GPT-4o,0.914)(Claude 3.5S,0.921)(Gemini 1.5P,0.896)(Llama3 70B,0.852)(Mixtral 8x22B,0.824)};
\addlegendentry{Governance}

\addplot[fill=red!40, draw=red!65]
  coordinates {(GPT-4o,0.887)(Claude 3.5S,0.898)(Gemini 1.5P,0.871)(Llama3 70B,0.827)(Mixtral 8x22B,0.798)};
\addlegendentry{Exec. Compensation}
\end{axis}
\end{tikzpicture}
\caption{Per-domain F1 across all five models under the hierarchical architecture. Financial metrics consistently achieve the highest scores due to standardized GAAP formats; executive compensation is the most challenging domain across all models.}
\label{fig:domain-f1}
\end{figure}

\subsection{Ablation Studies}

We conduct three ablation studies using the hierarchical architecture with Claude 3.5 Sonnet as the base configuration (F1 $= 0.929$, cost $= \$0.261$/doc).

\subsubsection{Semantic Caching}

We implement a semantic cache using embedding-based similarity matching (\texttt{text-embedding-3-small}, cosine threshold 0.95) on agent inputs. When a sufficiently similar input has been processed previously, the cached output is returned without an LLM call (Table~\ref{tab:caching}).

\begin{table}[ht]
\centering
\caption{Impact of semantic caching on the hierarchical architecture.}
\label{tab:caching}
\begin{tabular}{lcccc}
\toprule
\textbf{Cache Config} & \textbf{F1} & \textbf{Cost/doc} & \textbf{Lat (s)} & \textbf{Hit Rate} \\
\midrule
No cache          & 0.929 & \$0.261 & 46.2 & --- \\
Section-level     & 0.927 & \$0.198 & 34.8 & 24.1\% \\
Field-level       & 0.924 & \$0.171 & 29.3 & 38.7\% \\
Hybrid (adaptive) & 0.926 & \$0.182 & 31.4 & 31.4\% \\
\bottomrule
\end{tabular}
\end{table}

Field-level caching achieves the greatest cost reduction (34.5\%) with a modest F1 decrease of 0.005. The hybrid configuration, which caches field-level results for standard-format sections and disables caching for non-standard sections, achieves a favorable middle ground.

\subsubsection{Model Routing}

We evaluate a routing strategy where the supervisor agent assigns extraction tasks to different models based on estimated difficulty. Simple fields (binary governance indicators and standardized financial metrics) are routed to Mixtral 8x22B, while complex fields (compensation breakdowns and nuanced governance disclosures) are routed to Claude 3.5 Sonnet (Table~\ref{tab:routing}).

\begin{table}[ht]
\centering
\caption{Impact of model routing strategies.}
\label{tab:routing}
\begin{tabular}{lccc}
\toprule
\textbf{Routing Strategy} & \textbf{F1} & \textbf{Cost/doc} & \textbf{Lat (s)} \\
\midrule
All Claude 3.5 Sonnet           & 0.929 & \$0.261 & 46.2 \\
All GPT-4o                      & 0.921 & \$0.198 & 41.8 \\
All Mixtral 8x22B               & 0.843 & \$0.044 & 23.1 \\
2-tier (Claude + Mixtral)       & 0.912 & \$0.127 & 31.6 \\
3-tier (Claude + GPT + Mixtral) & 0.918 & \$0.143 & 33.2 \\
\bottomrule
\end{tabular}
\end{table}

The 2-tier routing strategy reduces cost by 51.3\% relative to all-Claude while retaining 98.2\% of the F1 score. The 3-tier strategy adds GPT-4o for medium-difficulty fields, recovering an additional 0.006 F1 at moderate cost increase.

\subsubsection{Adaptive Retry Strategies}

We compare retry strategies for handling low-confidence extractions in the hierarchical architecture (Table~\ref{tab:retry}).

\begin{table}[ht]
\centering
\caption{Retry strategy comparison.}
\label{tab:retry}
\begin{tabular}{lcccc}
\toprule
\textbf{Retry Strategy} & \textbf{F1} & \textbf{Cost/doc} & \textbf{Lat (s)} & \textbf{Retry Rate} \\
\midrule
No retry            & 0.908 & \$0.214 & 38.1 & 0\%    \\
Fixed retry (1x)    & 0.929 & \$0.261 & 46.2 & 14.2\% \\
Confidence-gated    & 0.926 & \$0.243 & 43.7 & 10.8\% \\
Escalation (model+) & 0.931 & \$0.258 & 45.1 & 12.1\% \\
Adaptive threshold  & 0.929 & \$0.247 & 44.0 & 11.6\% \\
\bottomrule
\end{tabular}
\end{table}

The escalation strategy, where failed extractions are retried with a more capable model, achieves the highest F1 (0.931) by directing difficult cases to stronger models. The adaptive threshold strategy, which raises the confidence threshold for fields with historically high error rates, matches the baseline F1 while reducing cost by 5.4\%.

\subsubsection{Combined Optimization}

Combining semantic caching (hybrid), model routing (2-tier) and adaptive retry yields a configuration we term \textbf{Hierarchical-Optimized} (Table~\ref{tab:combined}).

\begin{table}[ht]
\centering
\caption{Combined optimization: Hierarchical-Optimized configuration.}
\label{tab:combined}
\begin{tabular}{lccc}
\toprule
\textbf{Configuration} & \textbf{F1} & \textbf{Cost/doc} & \textbf{Lat (s)} \\
\midrule
Sequential baseline    & 0.903 & \$0.187 & 38.7 \\
Hierarchical baseline  & 0.929 & \$0.261 & 46.2 \\
Reflexive baseline     & 0.943 & \$0.430 & 74.1 \\
Hierarchical-Optimized & 0.924 & \$0.148 & 30.2 \\
\bottomrule
\end{tabular}
\end{table}

The Hierarchical-Optimized configuration achieves F1 of 0.924, recovering 89\% of the accuracy gap between sequential and reflexive baselines, at a cost of \$0.148/doc. This is only $1.15\times$ the sequential baseline cost and $0.57\times$ the hierarchical baseline cost, representing the most practically relevant finding of our study.

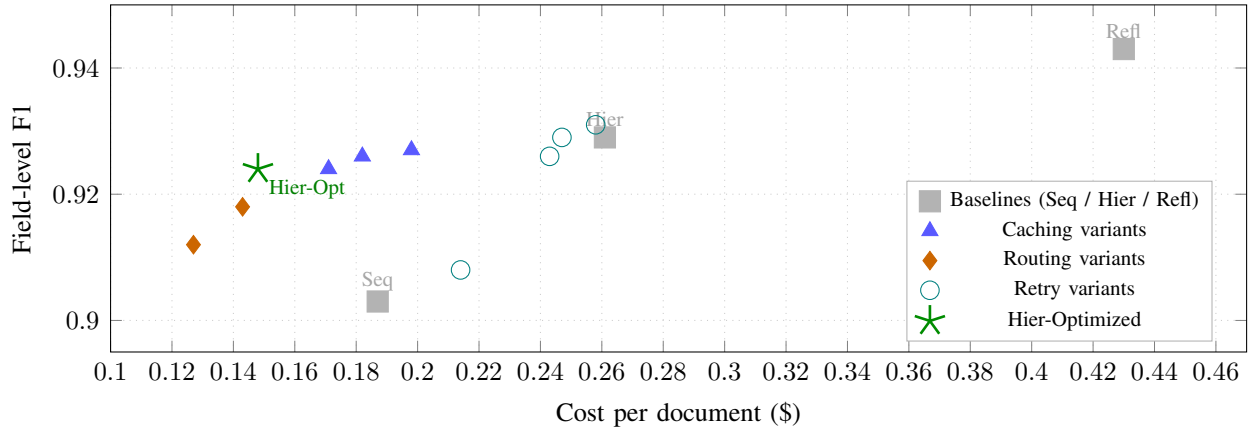
\begin{figure}[ht]
\centering
\begin{tikzpicture}
\begin{axis}[
  width=0.92\linewidth,
  height=6.2cm,
  xlabel={Cost per document (\$)},
  ylabel={Field-level F1},
  xlabel style={font=\small},
  ylabel style={font=\small},
  xticklabel style={font=\small},
  yticklabel style={font=\small},
  xmin=0.10, xmax=0.47,
  ymin=0.895, ymax=0.950,
  ymajorgrids=true, xmajorgrids=true,
  grid style={dotted, gray!40},
  legend pos=south east,
  legend style={font=\scriptsize, fill=white, draw=gray!60},
]
\addplot[only marks, mark=square*, mark size=4pt, color=gray!60]
  coordinates {(0.187,0.903)(0.261,0.929)(0.430,0.943)};
\addlegendentry{Baselines (Seq / Hier / Refl)}

\addplot[only marks, mark=triangle*, mark size=3.5pt, color=blue!65]
  coordinates {(0.198,0.927)(0.171,0.924)(0.182,0.926)};
\addlegendentry{Caching variants}

\addplot[only marks, mark=diamond*, mark size=3.5pt, color=orange!80!black]
  coordinates {(0.127,0.912)(0.143,0.918)};
\addlegendentry{Routing variants}

\addplot[only marks, mark=o, mark size=3.5pt, color=teal]
  coordinates {(0.214,0.908)(0.243,0.926)(0.258,0.931)(0.247,0.929)};
\addlegendentry{Retry variants}

\addplot[only marks, mark=star, mark size=6pt, color=green!55!black, line width=1pt]
  coordinates {(0.148,0.924)};
\addlegendentry{Hier-Optimized}

\node[font=\scriptsize, anchor=north west, text=green!50!black]
  at (axis cs:0.148,0.924) {Hier-Opt};
\node[font=\scriptsize, anchor=south, text=gray!70]
  at (axis cs:0.187,0.903) {Seq};
\node[font=\scriptsize, anchor=south, text=gray!70]
  at (axis cs:0.261,0.929) {Hier};
\node[font=\scriptsize, anchor=south, text=gray!70]
  at (axis cs:0.430,0.943) {Refl};
\end{axis}
\end{tikzpicture}
\caption{Ablation configurations plotted on the cost-F1 plane relative to the three baselines (gray squares). Routing variants achieve the largest cost reductions; the combined Hierarchical-Optimized configuration (green star) sits below baseline cost while recovering most of the accuracy gain.}
\label{fig:ablation-scatter}
\end{figure}

\subsection{Scaling Analysis}

We evaluate throughput-accuracy characteristics by varying the daily document processing volume from 1K to 100K under fixed compute budgets representative of production deployments (Table~\ref{tab:scaling}).

\begin{table}[ht]
\centering
\caption{Scaling characteristics (Claude 3.5 Sonnet, fixed 8-node cluster). F1 scores at varying daily throughput.}
\label{tab:scaling}
\begin{tabular}{lcccc}
\toprule
\textbf{Docs/Day} & \textbf{Seq. F1} & \textbf{Par. F1} & \textbf{Hier. F1} & \textbf{Refl. F1} \\
\midrule
1,000   & 0.903 & 0.914 & 0.929 & 0.943 \\
5,000   & 0.903 & 0.914 & 0.928 & 0.941 \\
10,000  & 0.901 & 0.913 & 0.926 & 0.937 \\
25,000  & 0.899 & 0.911 & 0.919 & 0.924 \\
50,000  & 0.894 & 0.906 & 0.907 & 0.898 \\
100,000 & 0.886 & 0.898 & 0.891 & 0.871 \\
\bottomrule
\end{tabular}
\end{table}

\textbf{Key Finding 4: Reflexive architecture degrades fastest at scale.} The reflexive architecture maintains its accuracy advantage up to approximately 25K docs/day but degrades sharply beyond that point. At 50K docs/day, it falls below the hierarchical architecture and at 100K docs/day it is the worst-performing architecture. This is because the iterative correction loops create queuing delays under high load; when timeout constraints are imposed, correction iterations are truncated, eliminating the architecture's primary advantage.

\textbf{Key Finding 5: Sequential is the most scale-resilient.} The sequential architecture shows the smallest absolute F1 degradation from 1K to 100K docs/day (0.017), owing to its deterministic execution profile and absence of coordination overhead. However, its absolute F1 remains below parallel and hierarchical at all tested scales up to 50K.

The architectural crossover point, where hierarchical F1 drops below parallel F1, occurs at approximately 75K docs/day on our test cluster. This threshold is infrastructure-dependent and will shift with compute scaling, but the qualitative pattern (reflexive degrades fastest, sequential degrades slowest) is robust.

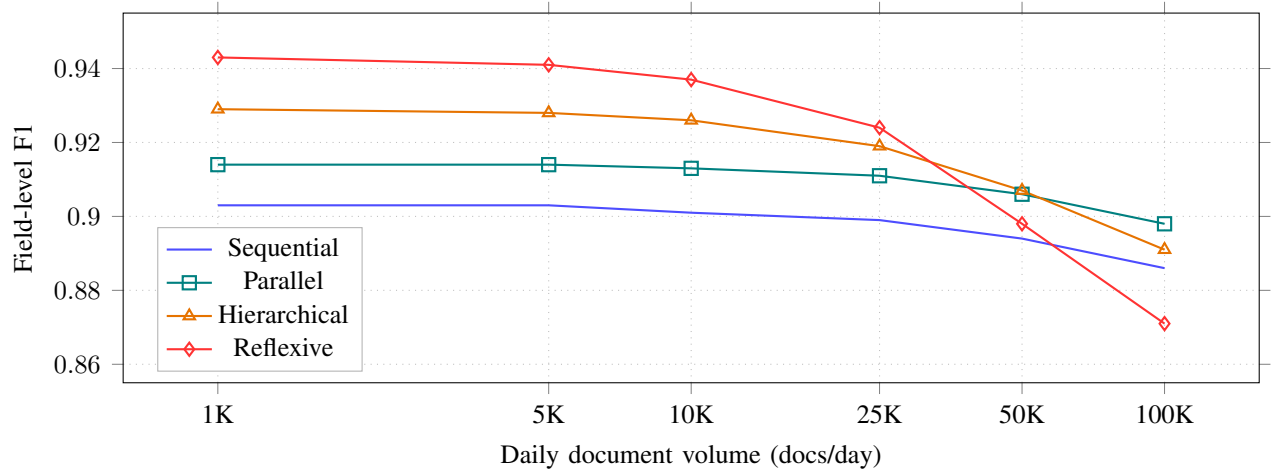
\begin{figure}[ht]
\centering
\begin{tikzpicture}
\begin{axis}[
  width=0.92\linewidth,
  height=6.5cm,
  xlabel={Daily document volume (docs/day)},
  ylabel={Field-level F1},
  xmode=log,
  log ticks with fixed point,
  xtick={1000,5000,10000,25000,50000,100000},
  xticklabels={1K,5K,10K,25K,50K,100K},
  xticklabel style={font=\small},
  yticklabel style={font=\small},
  xlabel style={font=\small},
  ylabel style={font=\small},
  ymin=0.855, ymax=0.955,
  ymajorgrids=true,
  xmajorgrids=true,
  grid style={dotted, gray!50},
  legend pos=south west,
  legend style={font=\small, fill=white, draw=gray!60},
  tick align=outside,
]
\addplot[color=blue!70, mark=circle, thick, mark size=2.5pt]
  coordinates {(1000,0.903)(5000,0.903)(10000,0.901)(25000,0.899)(50000,0.894)(100000,0.886)};
\addlegendentry{Sequential}

\addplot[color=teal, mark=square, thick, mark size=2.5pt]
  coordinates {(1000,0.914)(5000,0.914)(10000,0.913)(25000,0.911)(50000,0.906)(100000,0.898)};
\addlegendentry{Parallel}

\addplot[color=orange!90!black, mark=triangle, thick, mark size=2.5pt]
  coordinates {(1000,0.929)(5000,0.928)(10000,0.926)(25000,0.919)(50000,0.907)(100000,0.891)};
\addlegendentry{Hierarchical}

\addplot[color=red!80, mark=diamond, thick, mark size=2.5pt]
  coordinates {(1000,0.943)(5000,0.941)(10000,0.937)(25000,0.924)(50000,0.898)(100000,0.871)};
\addlegendentry{Reflexive}
\end{axis}
\end{tikzpicture}
\caption{F1 degradation as daily document volume scales from 1K to 100K. Reflexive accuracy collapses sharply beyond 25K docs/day due to timeout truncation of correction loops, while sequential degrades most gradually.}
\label{fig:scaling}
\end{figure}

\subsection{Latency Distribution Analysis}

Latency characteristics differ across architectures in ways not captured by median statistics alone (Table~\ref{tab:latency}).

\begin{table}[ht]
\centering
\caption{Latency percentiles in seconds (Claude 3.5 Sonnet, 10K docs).}
\label{tab:latency}
\begin{tabular}{lccc}
\toprule
\textbf{Architecture} & $\boldsymbol{p_{50}}$ & $\boldsymbol{p_{95}}$ & $\boldsymbol{p_{99}}$ \\
\midrule
Sequential   & 38.7  & 62.4  &  89.1 \\
Parallel     & 21.3  & 41.7  &  68.3 \\
Hierarchical & 46.2  & 78.3  & 124.6 \\
Reflexive    & 74.1  & 148.7 & 247.3 \\
\bottomrule
\end{tabular}
\end{table}

The reflexive architecture exhibits the widest latency distribution ($p_{99}/p_{50}$ ratio of $3.34\times$), reflecting the variable number of correction iterations. This makes it unsuitable for applications with strict latency SLAs unless combined with aggressive timeout policies, which, as shown in Section~\ref{sec:results}, degrade its accuracy advantage.

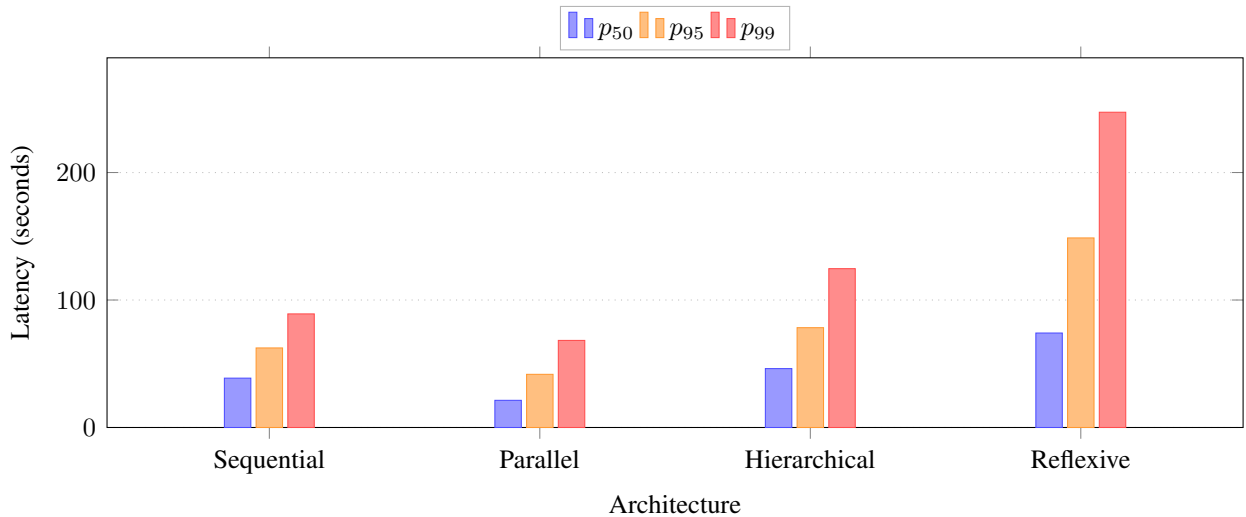
\begin{figure}[ht]
\centering
\begin{tikzpicture}
\begin{axis}[
  width=0.92\linewidth,
  height=6.5cm,
  ybar,
  bar width=10pt,
  xlabel={Architecture},
  ylabel={Latency (seconds)},
  xlabel style={font=\small},
  ylabel style={font=\small},
  xticklabel style={font=\small},
  yticklabel style={font=\small},
  symbolic x coords={Sequential, Parallel, Hierarchical, Reflexive},
  xtick=data,
  ymin=0, ymax=290,
  ymajorgrids=true,
  grid style={dotted, gray!50},
  legend style={font=\small, at={(0.5,1.02)}, anchor=south, legend columns=3, fill=white, draw=gray!60},
  enlarge x limits=0.2,
]
\addplot[fill=blue!40, draw=blue!70]
  coordinates {(Sequential,38.7)(Parallel,21.3)(Hierarchical,46.2)(Reflexive,74.1)};
\addlegendentry{$p_{50}$}

\addplot[fill=orange!50, draw=orange!80]
  coordinates {(Sequential,62.4)(Parallel,41.7)(Hierarchical,78.3)(Reflexive,148.7)};
\addlegendentry{$p_{95}$}

\addplot[fill=red!45, draw=red!70]
  coordinates {(Sequential,89.1)(Parallel,68.3)(Hierarchical,124.6)(Reflexive,247.3)};
\addlegendentry{$p_{99}$}
\end{axis}
\end{tikzpicture}
\caption{Latency percentile distribution ($p_{50}$, $p_{95}$, $p_{99}$) per architecture (Claude 3.5 Sonnet, 10K documents). The reflexive architecture shows the widest spread, with a $p_{99}/p_{50}$ ratio of $3.34\times$, making it unsuitable for strict SLA environments.}
\label{fig:latency-dist}
\end{figure}

\section{Analysis and Discussion}
\label{sec:discussion}

\subsection{Architecture Selection Guidelines}

Our results suggest the following decision framework for practitioners.

\textbf{Use Sequential when:} (a) the extraction task is simple (fewer than 10 field types); (b) documents are short and well-formatted; (c) cost is the primary constraint; or (d) processing volume exceeds 75K documents per day on constrained infrastructure. The sequential architecture's predictability and low overhead make it the pragmatic default.

\textbf{Use Parallel when:} (a) latency is critical (e.g., real-time processing pipelines); (b) extraction domains are naturally independent; or (c) the system must handle bursty workloads. The parallel architecture's ability to exploit multi-core and multi-GPU infrastructure directly translates input parallelism into latency reduction.

\textbf{Use Hierarchical when:} (a) maximum cost-efficiency at high accuracy is required; (b) the field set includes both simple and complex extraction targets; or (c) the system must operate at moderate scale (10K--50K docs/day) without accuracy degradation. The hierarchical architecture's adaptive routing and selective retry mechanisms make it the most production-suitable choice for most financial document processing scenarios.

\textbf{Use Reflexive when:} (a) extraction accuracy is paramount and cost is secondary (e.g., compliance-critical fields for regulatory submissions); (b) processing volume is low (under 10K docs/day); or (c) the output is consumed by downstream processes intolerant of errors. The reflexive architecture's self-correction mechanism is most valuable precisely when errors are most costly.

\begin{figure}[ht]
\centering
\begin{tikzpicture}[scale=1.05]

  \def\N{5}          
  \def\R{2.2}        

  \foreach \i/\lbl in {
    0/Accuracy,
    1/Speed,
    2/{Cost Eff.},
    3/{Scale Resilience},
    4/{Token Eff.}
  }{
    \pgfmathsetmacro{\angle}{90 - \i*360/\N}
    \draw[gray!40, thin] (0,0) -- (\angle:\R);
    \foreach \r in {0.55, 1.1, 1.65, 2.2}{
      \draw[gray!25, thin] (\angle:\r) -- ({90-(\i+1)*360/\N}:\r);
    }
    \node[font=\scriptsize, anchor={180+\angle}] at (\angle:{\R+0.38}) {\lbl};
  }

  \newcommand{\radarplot}[6]{%
    \def\col{#1}%
    \pgfmathsetmacro{\ax}{90}
    \pgfmathsetmacro{\bx}{90-72}
    \pgfmathsetmacro{\cx}{90-144}
    \pgfmathsetmacro{\dx}{90-216}
    \pgfmathsetmacro{\ex}{90-288}
    \fill[\col, opacity=0.15]
      (\ax:{#2*\R}) -- (\bx:{#3*\R}) -- (\cx:{#4*\R}) --
      (\dx:{#5*\R}) -- (\ex:{#6*\R}) -- cycle;
    \draw[\col, thick]
      (\ax:{#2*\R}) -- (\bx:{#3*\R}) -- (\cx:{#4*\R}) --
      (\dx:{#5*\R}) -- (\ex:{#6*\R}) -- cycle;
    \foreach \v/\ang in {#2/\ax, #3/\bx, #4/\cx, #5/\dx, #6/\ex}{
      \fill[\col] (\ang:{\v*\R}) circle (2pt);
    }%
  }

  \radarplot{blue!70}{0.60}{0.82}{1.00}{1.00}{0.60}
  \radarplot{teal}{0.72}{1.00}{0.84}{0.80}{0.40}
  \radarplot{orange!90!black}{0.86}{0.68}{0.57}{0.60}{1.00}
  \radarplot{red!75}{1.00}{0.00}{0.00}{0.30}{1.00}

  \matrix[below right, row sep=2pt, column sep=8pt, font=\scriptsize]
    at (2.4,-2.4) {
    \fill[blue!70] (0,0) circle (3pt); & \node{Sequential}; \\
    \fill[teal]    (0,0) circle (3pt); & \node{Parallel}; \\
    \fill[orange!90!black] (0,0) circle (3pt); & \node{Hierarchical}; \\
    \fill[red!75]  (0,0) circle (3pt); & \node{Reflexive}; \\
  };
\end{tikzpicture}
\caption{Radar chart comparing all four architectures across five normalized performance dimensions (higher = better on all axes; Speed and Cost Efficiency are inverted from their raw measures). Hierarchical dominates on token efficiency and balances accuracy and cost well. Reflexive leads on accuracy but scores zero on speed and cost efficiency. Parallel uniquely leads on speed. Sequential is the most cost-efficient and scale-resilient.}
\label{fig:radar}
\end{figure}
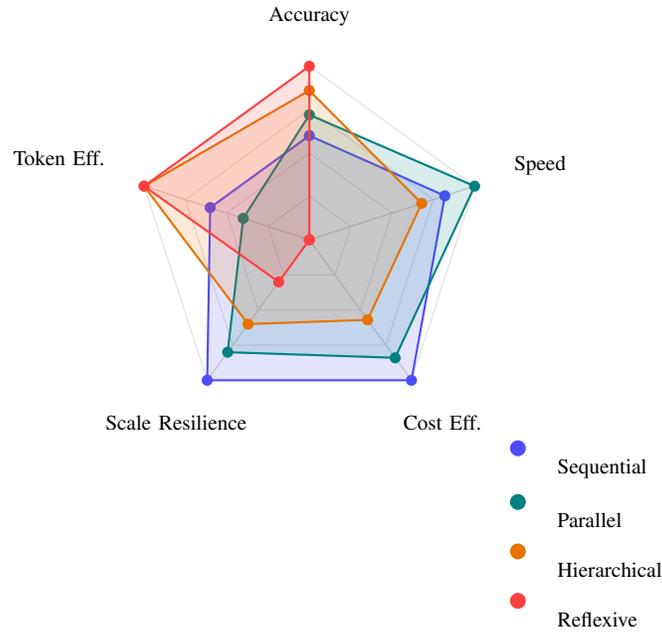

\subsection{Cost-Accuracy Pareto Frontier}

We construct the Pareto frontier across all tested configurations by plotting F1 against cost per document. The Hierarchical-Optimized configuration (\$0.148, F1 $= 0.924$) represents a particularly attractive knee point on the Pareto frontier, offering near-reflexive accuracy at near-sequential cost. For organizations willing to spend more, the hierarchical baseline with Claude 3.5 Sonnet (\$0.261, F1 $= 0.929$) offers the next significant quality improvement.

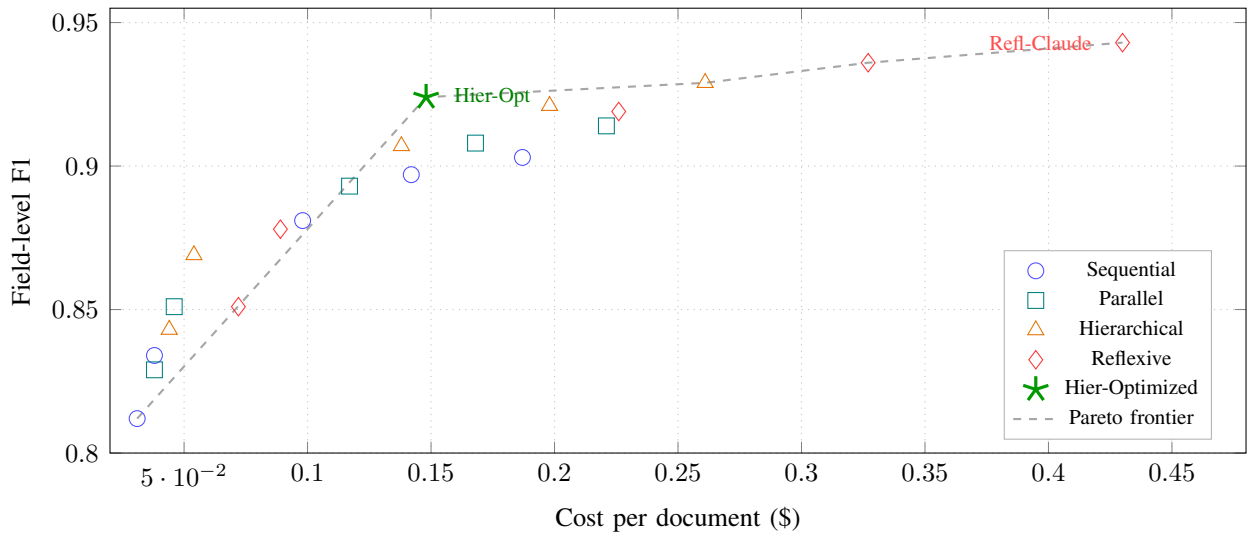
\begin{figure}[ht]
\centering
\begin{tikzpicture}
\begin{axis}[
  width=0.92\linewidth,
  height=7.5cm,
  xlabel={Cost per document (\$)},
  ylabel={Field-level F1},
  xlabel style={font=\small},
  ylabel style={font=\small},
  xticklabel style={font=\small},
  yticklabel style={font=\small},
  xmin=0.02, xmax=0.48,
  ymin=0.800, ymax=0.955,
  ymajorgrids=true,
  xmajorgrids=true,
  grid style={dotted, gray!50},
  legend pos=south east,
  legend style={font=\scriptsize, fill=white, draw=gray!60},
]

\addplot[only marks, mark=o, mark size=3pt, color=blue!70]
  coordinates {
    (0.031,0.812)  
    (0.038,0.834)  
    (0.098,0.881)  
    (0.142,0.897)  
    (0.187,0.903)  
  };
\addlegendentry{Sequential}

\addplot[only marks, mark=square, mark size=3pt, color=teal]
  coordinates {
    (0.038,0.829)
    (0.046,0.851)
    (0.117,0.893)
    (0.168,0.908)
    (0.221,0.914)
  };
\addlegendentry{Parallel}

\addplot[only marks, mark=triangle, mark size=3.5pt, color=orange!90!black]
  coordinates {
    (0.044,0.843)
    (0.054,0.869)
    (0.138,0.907)
    (0.198,0.921)
    (0.261,0.929)
  };
\addlegendentry{Hierarchical}

\addplot[only marks, mark=diamond, mark size=3.5pt, color=red!80]
  coordinates {
    (0.072,0.851)
    (0.089,0.878)
    (0.226,0.919)
    (0.327,0.936)
    (0.430,0.943)
  };
\addlegendentry{Reflexive}

\addplot[only marks, mark=star, mark size=5pt, color=green!60!black, line width=1.2pt]
  coordinates {(0.148,0.924)};
\addlegendentry{Hier-Optimized}

\addplot[dashed, thick, color=gray!70, no marks]
  coordinates {
    (0.031,0.812)
    (0.148,0.924)
    (0.261,0.929)
    (0.327,0.936)
    (0.430,0.943)
  };
\addlegendentry{Pareto frontier}

\node[font=\scriptsize, anchor=west, text=green!50!black]
  at (axis cs:0.155,0.924) {Hier-Opt};

\node[font=\scriptsize, anchor=east, text=red!70]
  at (axis cs:0.422,0.943) {Refl-Claude};

\end{axis}
\end{tikzpicture}
\caption{Cost-accuracy Pareto frontier across all architecture-model configurations. Each marker represents one architecture-model pair. The dashed line traces the Pareto frontier. The Hierarchical-Optimized point (green star) offers the best cost-efficiency knee, achieving F1 $= 0.924$ at only \$0.148/doc.}
\label{fig:pareto}
\end{figure}

\subsection{Failure Mode Analysis}

We identify and categorize 12 failure modes through manual analysis of 500 error cases (125 per architecture). Table~\ref{tab:failure-modes} reports the prevalence of the top failure modes by architecture.

\begin{table}[ht]
\centering
\caption{Failure mode prevalence (\% of errors attributable to each mode).}
\label{tab:failure-modes}
\small
\begin{tabular}{lcccc}
\toprule
\textbf{Failure Mode} & \textbf{Seq.} & \textbf{Par.} & \textbf{Hier.} & \textbf{Refl.} \\
\midrule
Cross-table reference failure     & 28.4 & 12.1 & 14.3 &  8.7 \\
Temporal confusion (FY vs. QTR)   & 18.2 & 17.8 & 12.1 &  6.4 \\
Unit/scale error (M vs. K vs. raw)& 14.7 & 14.2 & 10.8 &  4.2 \\
Proxy stmt. vs. 10-K confusion    &  9.3 &  6.8 &  7.2 &  5.1 \\
Restated figure extraction        &  8.1 &  9.4 &  8.7 &  7.8 \\
Compensation vesting misparse     &  7.6 &  8.3 &  9.1 &  8.4 \\
Context window truncation         &  5.8 & 14.7 &  6.2 &  3.8 \\
Agent coordination failure        &  0.0 &  8.1 & 12.4 & 14.2 \\
Hallucinated field value          &  4.9 &  4.3 &  3.8 &  2.1 \\
Ambiguous disclosure resolution   &  3.0 &  4.3 & 15.4 & 39.3 \\
\bottomrule
\end{tabular}
\end{table}

Several patterns emerge. First, cross-table reference failures, where a field's value requires linking data across multiple tables or sections, are the dominant failure mode for the sequential architecture (28.4\%) because information is processed in a fixed order that may not revisit earlier sections. The parallel and hierarchical architectures mitigate this through reconciliation and supervisor oversight, respectively, while the reflexive architecture's verification step catches most such errors.

Second, agent coordination failures (conflicts, deadlocks and message corruption) are absent in the sequential architecture but affect all multi-agent architectures. Notably, coordination failures increase from parallel (8.1\%) to hierarchical (12.4\%) to reflexive (14.2\%), reflecting the increasing complexity of inter-agent communication.

Third, the reflexive architecture's dominant failure mode is ambiguous disclosure resolution (39.3\% of its errors). When a filing genuinely contains ambiguous language, the reflexive architecture's correction loop can oscillate between interpretations across iterations, sometimes settling on a less likely reading. This ``overthinking'' failure mode is unique to iterative architectures.

\begin{figure}[ht]
\centering
\begin{tikzpicture}[font=\scriptsize]
  \def\cw{2.6cm}  
  \def\ch{0.52cm} 



  \newcommand{\hcell}[4]{%
    \pgfmathsetmacro{\intensity}{min(#3/40,1)}%
    \pgfmathsetmacro{\r}{0.15 + 0.85*(1-\intensity)}
    \pgfmathsetmacro{\g}{0.15 + 0.85*(1-\intensity)}%
    \pgfmathsetmacro{\b}{0.55 + 0.45*(1-\intensity)}%
    \pgfmathsetmacro{\r}{min(1, 0.2 + 1.6*\intensity)}%
    \pgfmathsetmacro{\g}{max(0, 0.85 - 1.2*\intensity)}%
    \pgfmathsetmacro{\b}{max(0, 0.85 - 1.2*\intensity)}%
    \fill[fill={rgb,1:red,\r;green,\g;blue,\b}]
      ({#1*\cw}, {-#2*\ch}) rectangle ({(#1+1)*\cw}, {-(#2+1)*\ch});%
    \node[anchor=center] at ({(#1+0.5)*\cw}, {-(#2+0.5)*\ch}) {#4};%
  }

  \node[anchor=center, font=\bfseries\scriptsize] at ({0.5*\cw},0.35) {Seq.};
  \node[anchor=center, font=\bfseries\scriptsize] at ({1.5*\cw},0.35) {Par.};
  \node[anchor=center, font=\bfseries\scriptsize] at ({2.5*\cw},0.35) {Hier.};
  \node[anchor=center, font=\bfseries\scriptsize] at ({3.5*\cw},0.35) {Refl.};

  \def\labels{{"Cross-table ref.","Temporal confusion","Unit/scale error","Proxy vs.\ 10-K","Restated figure","Vesting misparse","Ctx.\ truncation","Coordination fail","Hallucination","Ambiguous discl."}}
  \foreach \i in {0,...,9}{
    \pgfmathparse{\labels[\i]}\let\lbl\pgfmathresult
    \node[anchor=east, font=\scriptsize] at (0, {-(\i+0.5)*\ch}) {\lbl};
  }

  \hcell{0}{0}{28.4}{28.4}  \hcell{1}{0}{12.1}{12.1}  \hcell{2}{0}{14.3}{14.3}  \hcell{3}{0}{8.7}{8.7}
  \hcell{0}{1}{18.2}{18.2}  \hcell{1}{1}{17.8}{17.8}  \hcell{2}{1}{12.1}{12.1}  \hcell{3}{1}{6.4}{6.4}
  \hcell{0}{2}{14.7}{14.7}  \hcell{1}{2}{14.2}{14.2}  \hcell{2}{2}{10.8}{10.8}  \hcell{3}{2}{4.2}{4.2}
  \hcell{0}{3}{9.3}{9.3}    \hcell{1}{3}{6.8}{6.8}    \hcell{2}{3}{7.2}{7.2}    \hcell{3}{3}{5.1}{5.1}
  \hcell{0}{4}{8.1}{8.1}    \hcell{1}{4}{9.4}{9.4}    \hcell{2}{4}{8.7}{8.7}    \hcell{3}{4}{7.8}{7.8}
  \hcell{0}{5}{7.6}{7.6}    \hcell{1}{5}{8.3}{8.3}    \hcell{2}{5}{9.1}{9.1}    \hcell{3}{5}{8.4}{8.4}
  \hcell{0}{6}{5.8}{5.8}    \hcell{1}{6}{14.7}{14.7}  \hcell{2}{6}{6.2}{6.2}    \hcell{3}{6}{3.8}{3.8}
  \hcell{0}{7}{0.0}{0.0}    \hcell{1}{7}{8.1}{8.1}    \hcell{2}{7}{12.4}{12.4}  \hcell{3}{7}{14.2}{14.2}
  \hcell{0}{8}{4.9}{4.9}    \hcell{1}{8}{4.3}{4.3}    \hcell{2}{8}{3.8}{3.8}    \hcell{3}{8}{2.1}{2.1}
  \hcell{0}{9}{3.0}{3.0}    \hcell{1}{9}{4.3}{4.3}    \hcell{2}{9}{15.4}{15.4}  \hcell{3}{9}{39.3}{\textbf{39.3}}

  \draw[gray!50, thin] (0,0) grid[xstep=\cw, ystep=\ch] ({4*\cw},{-10*\ch});
  \draw[black, thick] (0,0) rectangle ({4*\cw},{-10*\ch});

  \node[font=\tiny] at ({4.3*\cw}, 0.1) {High};
  \foreach \v in {0,1,...,9}{
    \pgfmathsetmacro{\intensity}{\v/9}
    \pgfmathsetmacro{\r}{min(1, 0.2 + 1.6*\intensity)}
    \pgfmathsetmacro{\g}{max(0, 0.85 - 1.2*\intensity)}
    \pgfmathsetmacro{\b}{max(0, 0.85 - 1.2*\intensity)}
    \fill[fill={rgb,1:red,\r;green,\g;blue,\b}]
      ({4.15*\cw}, {-\v*\ch}) rectangle ({4.4*\cw},{-(\v+1)*\ch});
  }
  \node[font=\tiny] at ({4.3*\cw}, {-10*\ch-0.15}) {Low};
  \node[font=\tiny, rotate=90] at ({4.55*\cw},{-5*\ch}) {\% of errors};
\end{tikzpicture}
\caption{Failure mode heatmap showing the percentage of errors attributable to each failure category per architecture. Darker red indicates higher prevalence. The reflexive architecture's dominant failure (ambiguous disclosure resolution, 39.3\%) and the sequential architecture's cross-table reference failures (28.4\%) are clearly visible.}
\label{fig:heatmap}
\end{figure}
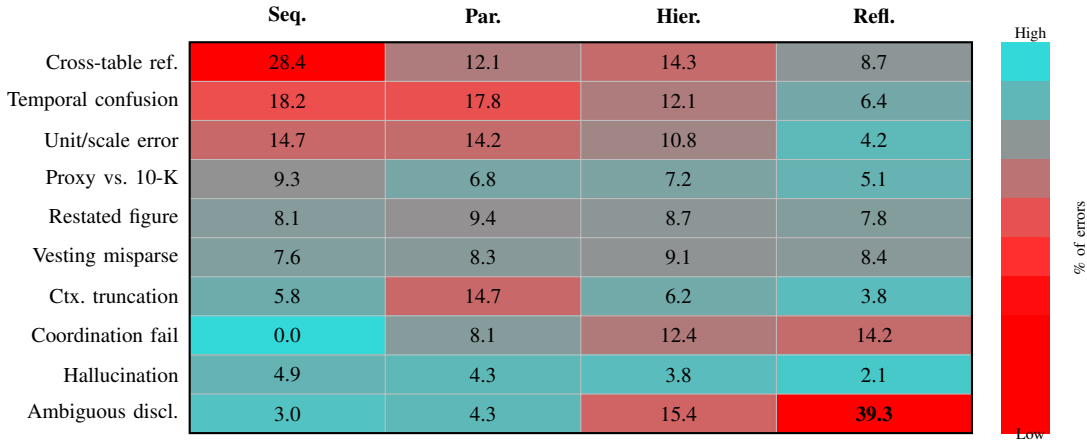

\subsection{Token Efficiency Analysis}

Token efficiency, the ratio of useful output information to total tokens consumed, reveals the computational overhead of each architecture (Table~\ref{tab:token-efficiency}).

\begin{table}[ht]
\centering
\caption{Token consumption breakdown (Claude 3.5 Sonnet, mean per document).}
\label{tab:token-efficiency}
\begin{tabular}{lrrrc}
\toprule
\textbf{Architecture} & \textbf{Input} & \textbf{Output} & \textbf{Total} & \textbf{Eff.} \\
\midrule
Sequential   & 142,340 &  3,820 & 146,160 & 2.61\% \\
Parallel     & 168,720 &  4,210 & 172,930 & 2.43\% \\
Hierarchical & 197,480 &  5,640 & 203,120 & 2.78\% \\
Reflexive    & 312,670 &  8,940 & 321,610 & 2.78\% \\
\bottomrule
\end{tabular}
\end{table}

Interestingly, hierarchical and reflexive architectures achieve the same token efficiency ratio (2.78\%) despite consuming vastly different total tokens. This is because both architectures generate proportionally more useful output per token: the hierarchical through targeted extraction and the reflexive through iterative refinement of outputs. The parallel architecture is the least token-efficient (2.43\%) due to overlapping context windows across branches that result in redundant input processing.

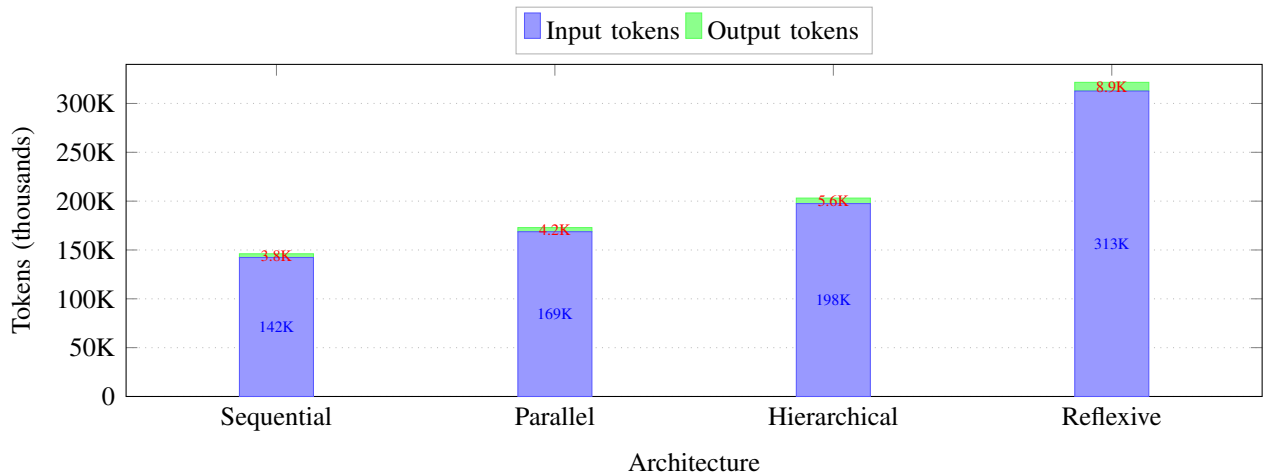
\begin{figure}[ht]
\centering
\begin{tikzpicture}
\begin{axis}[
  width=0.92\linewidth,
  height=6.0cm,
  ybar stacked,
  bar width=28pt,
  xlabel={Architecture},
  ylabel={Tokens (thousands)},
  xlabel style={font=\small},
  ylabel style={font=\small},
  xticklabel style={font=\small},
  yticklabel style={font=\small},
  symbolic x coords={Sequential, Parallel, Hierarchical, Reflexive},
  xtick=data,
  ymin=0, ymax=340,
  ytick={0,50,100,150,200,250,300},
  yticklabels={0,50K,100K,150K,200K,250K,300K},
  ymajorgrids=true,
  grid style={dotted, gray!50},
  legend style={font=\small, at={(0.5,1.03)}, anchor=south, legend columns=2, fill=white, draw=gray!60},
  enlarge x limits=0.18,
  nodes near coords,
  nodes near coords style={font=\tiny, /pgf/number format/fixed, /pgf/number format/precision=1},
  point meta=explicit symbolic,
]
\addplot+[fill=blue!40, draw=blue!65, nodes near coords, point meta=explicit symbolic]
  coordinates {
    (Sequential,142.3)  [142K]
    (Parallel,168.7)    [169K]
    (Hierarchical,197.5)[198K]
    (Reflexive,312.7)   [313K]
  };
\addlegendentry{Input tokens}

\addplot+[fill=green!45, draw=green!65, nodes near coords, point meta=explicit symbolic]
  coordinates {
    (Sequential,3.8)  [3.8K]
    (Parallel,4.2)    [4.2K]
    (Hierarchical,5.6)[5.6K]
    (Reflexive,8.9)   [8.9K]
  };
\addlegendentry{Output tokens}
\end{axis}
\end{tikzpicture}
\caption{Stacked token consumption per architecture (input + output, mean per document, Claude 3.5 Sonnet). Output tokens are a small fraction in all cases. The reflexive architecture consumes $2.2\times$ more input tokens than sequential, explaining the cost differential despite identical token efficiency ratios.}
\label{fig:token-stack}
\end{figure}

\subsection{Limitations}

Our study has several limitations. First, our dataset is restricted to SEC filings in English; results may not generalize to other jurisdictions (e.g., IFRS filings) or languages. Second, our cost analysis uses API pricing as of January 2025 and will require updating as model pricing evolves. The directional findings (reflexive is most expensive, sequential is cheapest) are likely stable, but the specific cost ratios will shift. Third, our evaluation uses a fixed set of 25 field types; performance on novel field types not represented in our prompt engineering may differ. Fourth, we evaluate individual model versions and do not account for model degradation over time \cite{chen2024chatgpt}, which may affect production systems using API-served models. Fifth, our ground truth annotations, while produced by credentialed professionals, may contain systematic biases in ambiguous cases that affect apparent accuracy of different architectures.

\section{Conclusion and Future Work}
\label{sec:conclusion}

We have presented a comprehensive benchmark of four multi-agent LLM architectures for financial document processing, evaluated across five models, 25 extraction field types and 10,000 SEC filings. Our findings provide three actionable conclusions for practitioners.

\begin{enumerate}
  \item \textbf{The hierarchical architecture offers the best cost-accuracy tradeoff} for production financial document processing, achieving 98.5\% of the best-observed F1 at 60.7\% of the cost. When combined with semantic caching, model routing and adaptive retries, the optimized hierarchical configuration recovers 89\% of the reflexive architecture's accuracy gains at only $1.15\times$ the sequential baseline cost.

  \item \textbf{Architecture choice interacts with scale in non-obvious ways.} The reflexive architecture, which is the best performer at low volume, becomes the worst performer above 50K documents per day due to queuing-induced timeout truncation. Production architects must consider target scale in their architecture selection.

  \item \textbf{Failure modes are architecture-specific} and often represent the dark side of each architecture's strengths. The reflexive architecture's iterative correction, its primary advantage, also generates its dominant failure mode (oscillating ambiguity resolution). Practitioners should implement architecture-specific monitoring to detect these failure patterns.
\end{enumerate}

\subsection{Future Directions}

Several directions merit future investigation. First, \textbf{dynamic architecture switching}, routing individual documents to different architectures based on estimated complexity, could combine the efficiency of sequential processing for simple documents with the accuracy of reflexive processing for complex ones. Preliminary experiments suggest this could achieve F1 $> 0.935$ at cost below \$0.180/doc, but robust complexity estimation remains an open challenge.

Second, \textbf{fine-tuned specialist models} for specific extraction domains (e.g., a compensation-extraction model distilled from GPT-4o outputs) could dramatically reduce the cost of high-accuracy extraction. Our model routing ablation suggests that task-specific smaller models can handle 60--70\% of extraction tasks without quality loss.

Third, \textbf{streaming architectures} that process filings incrementally as they are published, extracting from early sections while later sections are still being parsed, could reduce effective latency by 40--60\% for long documents such as 10-K filings.

Fourth, \textbf{cross-document reasoning}, where extraction from one filing is informed by the same company's prior filings, could address temporal confusion errors and enable trend extraction, which is not possible with document-independent processing.

Finally, \textbf{formal verification} of financial identity constraints (e.g., balance sheet equations) as hard post-conditions could provide provable correctness guarantees for a subset of extracted fields, complementing the probabilistic assurances of LLM-based verification.

\bibliographystyle{IEEEtran}

\end{document}